\documentclass[manuscript]{acmart}
\AtBeginDocument{%
  }

\setcopyright{acmlicensed}
\copyrightyear{2026}
\acmYear{2026}
\acmDOI{XXXXXXX.XXXXXXX}
\acmConference[Conference acronym 'XX]{Make sure to enter the correct
  conference title from your rights confirmation email}{June 03--05,
  2026}{Woodstock, NY}
\acmISBN{978-1-4503-XXXX-X/2018/06}

\usepackage{multirow} 

\begin{document}




\title{Self-Evolving Cognitive Framework via Causal World Modeling for Embodied Scientific Intelligence}

\author{Yi Yu}
\authornote{Corresponding author. Email: yiyu@hiroshima-u.ac.jp}
\affiliation{%
  \institution{Graduate School of Advanced Science and Engineering, Hiroshima University}
  \city{Hiroshima}
  \country{Japan}
}
\email{yiyu@hiroshima-u.ac.jp}

\author{Tetsunari Inamura}
\affiliation{%
  \institution{Advanced Intelligence and Robotics Research Center, Brain Science Institute, Tamagawa University}
  \city{Tokyo}
  \country{Japan}
}
\email{inamura@ieee.org}
\renewcommand{\shortauthors}{Yu et al.}


\begin{abstract}

Current embodied world models are primarily optimized for predictive objectives, limiting their ability to generalize under distribution shifts and reason systematically about unseen situations and hypothetical interventions.
We argue that embodied intelligence should move beyond predictive world modeling toward self-evolving cognitive systems that continually construct and refine internal causal representations through interaction with the environment.
To this end, we propose a self-evolving cognitive framework via causal world modeling for embodied scientific intelligence, which integrates three complementary components: causal world modeling, intervention-driven causal reasoning, and continual cognitive refinement. The proposed framework continuously revises and expands its internal causal world model through causal discovery, intervention-driven feedback, and counterfactual reasoning, supporting continual cognitive refinement and enabling cognition itself to evolve over time. 
Furthermore, we reinterpret embodied interaction not merely as a means of trajectory optimization, but as an epistemic process for causal hypothesis generation, intervention-driven experimentation, and continual knowledge acquisition. This work provides a conceptual and theoretical foundation for a transition from predictive intelligence toward epistemic intelligence, in which intelligence emerges through the continual construction, revision, and refinement of causal world models via interaction with the environment. Accordingly,  an intervention-driven causal-epistemic benchmarking paradigm is suggested for evaluating self-evolving embodied scientific intelligence.

\end{abstract}




\maketitle


\section{Introduction}

Embodied intelligence has evolved from early symbolic reasoning and classical planning approaches~\cite{Fikes1971STRIPS,Kaelbling2011HTN} to data-driven perception--action learning methods~\cite{Brohan2022RT1}, and more recently toward large-scale multimodal agents and vision--language--action (VLA) models that integrate perception, language understanding, and action generation within a unified framework~\cite{Reed2022Gato,Brohan2023RT2,Driess2023PaLME}. These advances have significantly enhanced the capabilities of embodied agents across a wide range of robotic manipulation and long-horizon task execution scenarios in increasingly complex environments~\cite{Brohan2023RT2,Driess2023PaLME,Kim2024OpenVLA}.

In parallel, world models have emerged as a fundamental paradigm for model-based reinforcement learning by learning latent environmental dynamics for prediction and planning~\cite{Ha2018WorldModels,Hafner2020Dreamer,Schrittwieser2020MuZero,Ding2025}. By constructing compact latent representations of environmental dynamics, world models provide internal simulators that support efficient planning and sequential decision making~\cite{Ha2018WorldModels,Hafner2020Dreamer,Schrittwieser2020MuZero}. Despite their remarkable empirical success, existing world models are primarily optimized for predictive or correlation-based learning objectives and generally capture statistical regularities from observational trajectories rather than explicitly modeling the causal mechanisms governing environmental dynamics~\cite{Ha2018WorldModels,Hafner2023DreamerV3,Ding2025,Pearl2009Causality,Scholkopf2021}. Consequently, their ability to support intervention-aware reasoning, counterfactual inference, and continual refinement of internal representations through interaction remains fundamentally limited~\cite{Pearl2009Causality,Scholkopf2021,Parisi2019ContinualLearning,DeLange2021ContinualSurvey}.
In the context of embodied cognition, intelligent agents require capabilities beyond passive prediction. Rather than merely forecasting future observations, they should reason about intervention outcomes, evaluate counterfactual scenarios, explain observed behaviors, and continually refine internal knowledge through interaction with the environment.

Current embodied AI systems nevertheless exhibit three fundamental limitations. First, existing world models primarily capture observational regularities instead of explicitly modeling causal mechanisms, making it difficult to distinguish causal effects from statistical correlations~\cite{Pearl2009Causality,Scholkopf2021,Zeng2023CausalRLsurvey}. Second, although recent advances in causal reinforcement learning improve intervention-aware reasoning, most existing approaches assume fixed causal structures throughout learning and therefore cannot continually revise causal knowledge through interaction~\cite{Yang2023CausalWorldModelRL,Zeng2023CausalRLsurvey,Buesing2019CounterfactualRL,Ke2022CausalCounterfactualRL}. Third, continual learning methods mainly focus on parameter adaptation and catastrophic forgetting while lacking explicit mechanisms for continual refinement of internal causal representations~\cite{Parisi2019ContinualLearning,DeLange2021ContinualSurvey,Bousmalis2023RoboCat}.

These limitations reveal a fundamental gap between predictive world modeling and causal cognitive modeling for embodied scientific intelligence. Rather than treating cognition as predictive adaptation alone, embodied agents should continually construct, revise, and refine internal causal world models that explain environmental dynamics and support systematic reasoning under changing conditions~\cite{Pearl2009Causality,Peters2017Elements}.

Motivated by these observations, we propose a self-evolving cognitive framework via causal world modeling for embodied scientific intelligence. The proposed framework integrates causal world modeling, intervention-driven causal reasoning, and continual cognitive refinement into a unified cognitive process through which embodied agents progressively improve their internal understanding of the environment via interaction and environmental feedback. Rather than functioning as independent components, these processes continuously co-evolve through interaction-driven feedback, enabling causal knowledge itself to evolve over time.

Furthermore, we reinterpret embodied interaction and model-based imagination not merely as mechanisms for trajectory optimization, but as epistemic processes for causal hypothesis generation, intervention-driven experimentation, and continual knowledge acquisition. Through interaction with both the physical environment and internally imagined causal rollouts, embodied agents can evaluate hypothetical interventions, discover latent causal mechanisms, and iteratively revise their internal causal world models without relying solely on costly physical experimentation. Embodied intelligence emerges through the continual construction, revision, and refinement of causal world models via interaction and imagination, suggesting a transition from predictive intelligence toward epistemic intelligence. This transition also calls for a corresponding rethinking of evaluation paradigms. Conventional embodied AI benchmarks primarily emphasize task completion or trajectory prediction under relatively fixed environments, providing limited insight into the ability of an agent to acquire, revise, and maintain causal knowledge through continual interaction. We therefore argue that embodied intelligence should additionally be evaluated according to its capacity for intervention-aware reasoning, counterfactual consistency, causal model refinement, and continual epistemic accumulation. This conceptualization motivates an intervention-driven causal-epistemic benchmarking paradigm that complements the proposed self-evolving cognitive framework and provides a principled foundation for evaluating self-evolving embodied scientific intelligence.

\section{Related Work}

In this section, we briefly review representative research directions related to our work, discuss their limitations, and clarify how the proposed self-evolving cognitive framework differs from existing approaches.

\subsection{World Models for Embodied Intelligence}

World models have become a fundamental paradigm in model-based reinforcement learning by learning latent representations of environmental dynamics for prediction and planning~\cite{Ha2018WorldModels,Hafner2019PlaNet,Hafner2020Dreamer,Schrittwieser2020MuZero,Ding2025}. Recent advances have further extended world modeling to embodied intelligence, enabling robotic manipulation, long-horizon planning, multimodal decision making, and imagination-based control in complex environments~\cite{Brohan2023RT2,Driess2023PaLME,Kim2024OpenVLA,Hafner2023DreamerV3}.

Despite their strong empirical performance, existing world models remain predominantly predictive in nature. They are typically trained on observational trajectories and learn statistical regularities of dynamics without explicitly disentangling the underlying causal mechanisms governing environmental transitions~\cite{Pearl2009Causality,Scholkopf2021}. As a result, their generalization under interventions, structural changes, and out-of-distribution conditions remains limited.

\subsection{Causal Reinforcement Learning and Counterfactual Reasoning}

Causal reinforcement learning integrates structural causal models into sequential decision making to improve robustness, interpretability, and out-of-distribution generalization~\cite{Pearl2009Causality,Peters2017Elements,Zeng2023CausalRLsurvey,Yang2023CausalWorldModelRL}. Existing work explores causal discovery, intervention-aware policy learning, and counterfactual reasoning to evaluate hypothetical action outcomes beyond observational correlations~\cite{Buesing2019CounterfactualRL,Ke2022CausalCounterfactualRL}.

These approaches highlight the benefits of incorporating causal structure into embodied decision making. However, they typically assume a stationary causal structure and focus on leveraging, rather than continually revising, causal knowledge during learning. This limits their applicability in open-ended environments where causal mechanisms may be initially misspecified or subject to change.

\subsection{Continual Learning and Self-Improving Embodied Agents}

Continual learning studies how agents accumulate knowledge over sequential experiences while mitigating catastrophic forgetting~\cite{Parisi2019ContinualLearning,DeLange2021ContinualSurvey}. Existing approaches primarily focus on parameter-level adaptation via replay, regularization, or architectural expansion, enabling agents to learn new tasks without overwriting previously acquired knowledge.

Recent robotic foundation models and self-improving embodied agents further extend this paradigm through large-scale interaction, self-supervised experience accumulation, and autonomous skill refinement~\cite{Bousmalis2023RoboCat,Ju2026EmbodiSkill}. However, these methods mainly emphasize behavioral adaptation and skill acquisition, rather than continual refinement of the underlying causal mechanisms that explain environmental dynamics.

\subsection{Causal Representation Learning}

Causal representation learning aims to discover latent variables corresponding to the underlying causal factors that generate observations~\cite{Scholkopf2021,Peters2017Elements}. By identifying causally meaningful representations, these methods improve robustness, invariance, and transferability beyond purely statistical feature learning.

Although this line of work provides a strong theoretical foundation for uncovering latent causal structure, most existing approaches are developed under passive observational settings. They rarely incorporate active intervention, embodied interaction, or continual refinement of causal structure through environmental feedback, leaving causal learning in embodied settings largely unexplored.

\subsection{Position of This Work}

The above research directions address complementary but fragmented aspects of embodied intelligence. World models focus on predictive dynamics without explicit causal structure. Causal reinforcement learning introduces intervention-aware reasoning but typically assumes fixed causal graphs. Continual learning enables lifelong adaptation primarily at the parameter level rather than at the level of causal mechanism refinement. Causal representation learning provides a foundation for latent causal discovery but is largely developed under passive observational settings.

Encouraged by these constraints, this work proposes a unified view of embodied cognition as a self-evolving causal world modeling process. Instead of treating prediction, causal reasoning, continual adaptation, and representation learning as separate components, we integrate them into a single cognitive framework in which embodied agents continuously construct, evaluate, revise, and refine internal causal world models through interaction, intervention, counterfactual reasoning, and environmental feedback.
More fundamentally, embodied intelligence can be reframed  as a process of continual causal knowledge construction rather than purely predictive adaptation or policy optimization. In this view, agents function as epistemic systems that iteratively generate hypotheses, intervene in the environment, evaluate causal evidence, and update internal world models accordingly.

By unifying causal world modeling, intervention-driven reasoning, and continual cognitive refinement within a self-evolving framework, this work establishes a conceptual bridge between embodied intelligence and scientific inquiry. In particular, it suggests that embodied agents can be understood as autonomous causal learners operating within a scientific-like epistemic loop. Accordingly, we propose an intervention-driven causal-epistemic benchmarking paradigm for evaluating self-evolving embodied scientific intelligence.

\section{Causal Foundations of Self-Evolving Cognitive Framework}

\begin{figure}[t]
    \centering
    \includegraphics[width=1\linewidth,
    height=8cm]{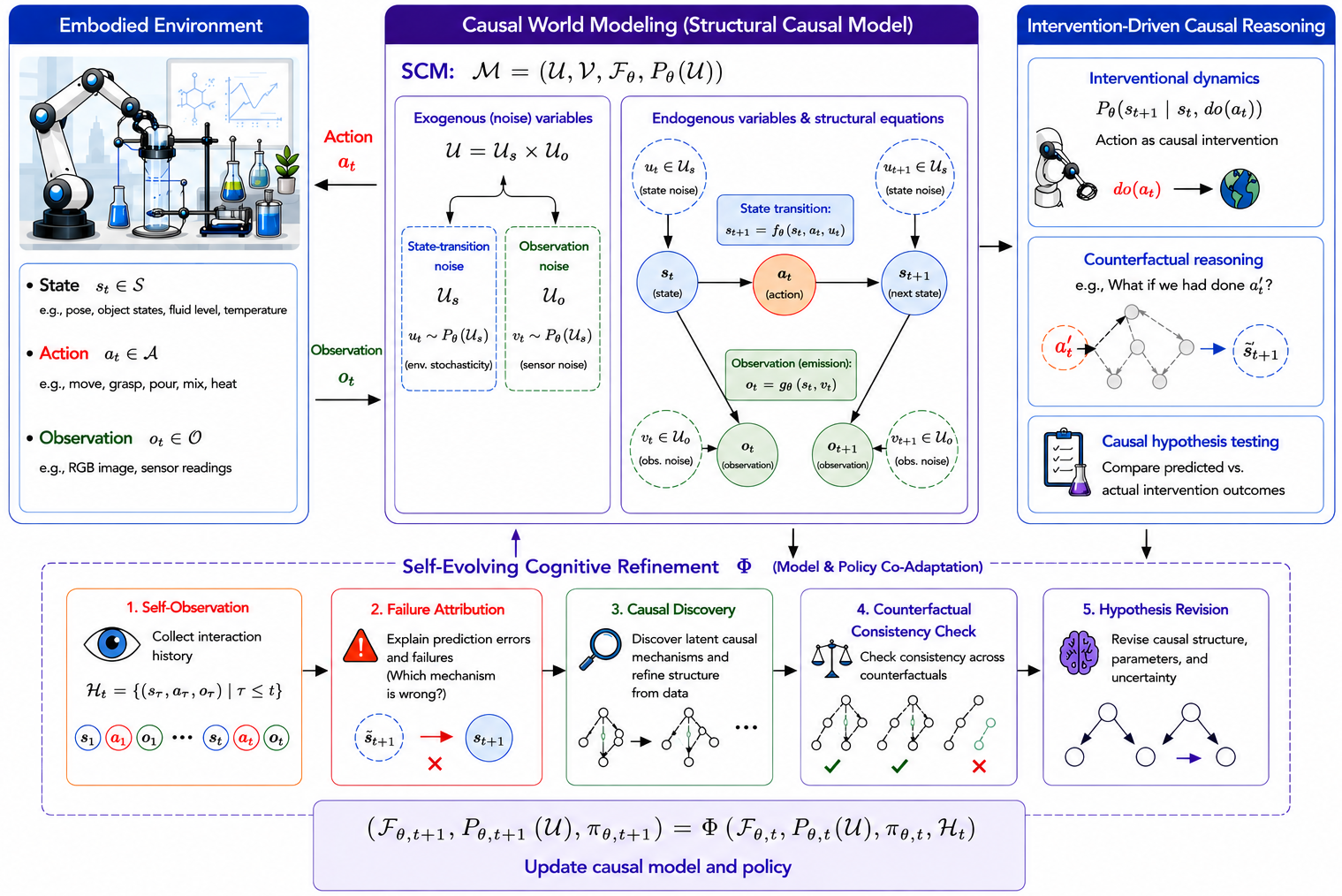}
    \caption{Self-Evolving Cognitive Framework  via Causal World Modeling for Embodied Scientific Intelligence.}
    \label{fig:self}
\end{figure}

To formalize the proposed self-evolving cognitive framework via causal world modeling, we first present an overview of the underlying cognitive process through which embodied agents construct, reason over, and continually refine their internal causal representations of the environment (Figure~\ref{fig:self}).
Within this framework, internal causal representations are continuously updated through interaction, intervention, counterfactual reasoning, and environmental feedback. More fundamentally, embodied cognition is viewed as the continual self-evolution of an internal causal world model driven by interaction, intervention, and causal discovery, rather than merely as predictive adaptation through policy optimization.
This self-evolving cognitive process enables embodied agents not only to predict future states but also to reason about latent causal mechanisms, evaluate hypothetical interventions, and progressively acquire more accurate causal knowledge of their environments through continual experience (see Table~\ref{tab:notation} for a summary of notations used in this framework).

\begin{table}[t]
\centering
\small
\caption{Summary of key notations used in the causal foundations of self-evolving cognitive framework.}
\label{tab:notation}
\begin{tabular}{|c|p{10.5cm}|}
\hline
\textbf{Notation} & \textbf{Meaning} \\
\hline

$\mathcal{M}$ & SCM representing environment dynamics \\
\hline

$\mathcal{M}_t$ & Time-indexed SCM representing the evolving causal world model at time $t$ \\
\hline

$\mathcal{U}$ & Exogenous variables capturing environmental stochasticity and unobserved external causal factors \\
\hline

$\mathcal{V}$ & Endogenous variables including state $s_t$, action $a_t$, and observation $o_t$ \\
\hline

$\mathcal{F}_{\theta}$ & Structural equations defining causal mechanisms in the static SCM \\
\hline

$\mathcal{F}_{\theta,t}$ & Time-evolving structural mechanisms in the self-evolving causal world model \\
\hline

$P_{\theta}(\mathcal{U})$ & Distribution over exogenous variables in the static SCM \\
\hline

$P_{\theta,t}(\mathcal{U})$ & Time-varying distribution over exogenous variables in the evolving model \\
\hline

$\theta$ & Learnable parameters governing structural mechanisms and noise distributions \\
\hline

$\pi_{\theta,t}$ & Time-dependent policy for action selection conditioned on interaction history $\mathcal{H}_t$ \\
\hline

$\mathcal{H}_t$ &
Interaction history $\{(s_{\tau},a_{\tau},o_{\tau})\mid \tau\le t\}$ \\
\hline

$\Phi$ & Structural evolution operator updating causal mechanisms, exogenous distribution, and policy based on interaction history \\
\hline

$do(a_t)$ & Causal intervention representing an externally imposed action in the SCM framework \\
\hline

\end{tabular}
\end{table}

\textbf{ Structural causal model.} From a causal perspective, we model an embodied environment as a structural causal process, where system dynamics are generated by underlying latent mechanisms. Let 
\begin{equation}
\mathcal{M} = (\mathcal{U}, \mathcal{V}, \mathcal{F}_{\theta}, P_{\theta}(\mathcal{U}))   \label{eq:scm}
\end{equation}
denote a structural causal model (SCM)~\cite{Pearl2009Causality}, where $\mathcal{U}$ denotes exogenous (unobserved) noise variables and $P_{\theta}(\mathcal{U})$ is its distribution, $\mathcal{V}$ denotes endogenous variables, $\mathcal{F}_{\theta} = \{ f_{\theta}, g_{\theta} \}$ denotes a set of structural equations specifying how each variable in $\mathcal{V}$ is generated from its causal parents and corresponding exogenous noise, and $\theta$ is a set of parameters. We further decompose the exogenous space into two independent components via a Cartesian product as $\mathcal{U} = \mathcal{U}_s \times  \mathcal{U}_o$, where $\mathcal{U}_s$ represents state-transition noise space, $\mathcal{U}_o$ represents observation noise space, and $P_{\theta}(\mathcal{U}) = P_{\theta}(\mathcal{U}_s) P_{\theta}(\mathcal{U}_o)$. 
This decomposition explicitly separates environmental uncertainty from perceptual uncertainty, providing a principled causal basis for intervention, counterfactual reasoning, and causal hypothesis testing in embodied environments.

We consider endogenous variables such as state variable $s_t \in \mathcal{S}$, action variable $a_t \in \mathcal{A}$, and observation variable $o_t \in \mathcal{O}$. 
Let $u_t \in \mathcal{U}_s$ ($u_t \sim P_{\theta}(\mathcal{U}_s)$) denote unobserved environmental stochasticity and latent factors affecting system dynamics, and $v_t \in \mathcal{U}_o$ ($v_t \sim P_{\theta}(\mathcal{U}_o)$) represent sensor noise and observation uncertainty.
The environment evolves according to the structural equation 
\begin{equation}
s_{t+1} = f_{\theta}(s_t, a_t, u_t),
\end{equation}
while observations are generated via an emission process
\begin{equation}
o_t = g_{\theta}(s_t, v_t).
\end{equation}

The embodied agent acts as an intervention generator, while the environment is a causal response system governed by latent mechanisms and exogenous disturbances.
Interventions correspond to manipulations of the data-generating process, whereas counterfactual reasoning evaluates alternative realizations of the same SCM under hypothetical interventions.

\textbf{From observational to causal world models.}
Standard world models implicitly approximate observational transition distributions $P_\theta(s_{t+1} \mid s_t, a_t)$~\cite{Ha2018WorldModels, Hafner2020Dreamer, Hafner2019PlaNet}, which capture statistical associations induced by observed data but do not explicitly distinguish observational correlations from interventional causal effects~\cite{Pearl2009Causality}. In contrast, the proposed causal decomposition explicitly separates environmental uncertainty from perceptual uncertainty, providing a principled causal foundation for intervention, counterfactual reasoning, and causal hypothesis testing in embodied environments. An intervention is defined as an action executed by the embodied agent that actively modifies the state of the environment. Formally, we model the action $a_t$ as a causal intervention $do(a_t)$ within a structural causal model of the environment. Under this interpretation, state transitions are governed by intervention-conditioned structural equations, thereby reflecting the causal effects of interventions rather than passive observations. Exogenous disturbances are treated as non-intervened environmental factors independent of agent actions.

We define a causal world model as a model that approximates interventional dynamics induced by an underlying structural causal model. Specifically, we consider interventional dynamics of the form $P_\theta(s_{t+1} \mid s_t, do(a_t))$, where actions are interpreted as causal interventions rather than passive inputs. Unlike observational transitions, interventional dynamics characterize the causal consequences of actions under explicit interventions.

\textbf{Self-evolving causal world modeling.}
To support open-ended embodied intelligence, we define a self-evolving causal world model as a causal world model equipped with a structural evolution mechanism that allows continual refinement of its internal causal representations through environmental feedback. A self-evolving causal world model is defined as a pair $(\mathcal{M}_t, \Phi)$,
where $\mathcal{M}_t$ is a time-indexed SCM and $\Phi$ is a structural evolution operator.
We formalize this by introducing a time-indexed extension of the structural causal model

\begin{equation}
\mathcal{M}_t
=
(\mathcal{U}, \mathcal{V}, \mathcal{F}_{\theta,t}, P_{\theta,t}(\mathcal{U})),
\end{equation}
alongside an adaptive policy $\pi_{\theta,t}$ governing action selection. The model and the policy are continuously updated based on interaction history
$\mathcal{H}_t = \{ (s_{\tau}, a_{\tau}, o_{\tau}) \mid \tau \le t \}$, by the following equation:
\begin{equation}
(\mathcal{F}_{\theta,t+1}, P_{\theta,t+1}(\mathcal{U}), \pi_{\theta,t+1}) = \Phi(\mathcal{F}_{\theta,t}, P_{\theta,t}(\mathcal{U}), \pi_{\theta,t}, \mathcal{H}_t),   
\end{equation}
where $\Phi$ induces updates driven by self-observation, failure attribution, intervention outcomes, counterfactual consistency checking, and causal hypothesis revision. Unlike static causal world models, a self-evolving causal world model continuously revises its internal causal structure, uncertainty estimates, and behavioral strategies through embodied interaction, enabling continual refinement of predictive accuracy, causal identifiability, and intervention-driven reasoning in open-ended and non-stationary environments.

To clarify the difference between prior work and the proposed framework, we formalize different paradigms in terms of constraints on admissible model transformations.

\textit{(1) Parameter-only adaptation (conventional learning).}
Conventional learning assumes a fixed causal structure and updates only parameters, typically written as
$\theta_{t+1} = \Phi(\theta_t, \mathcal{H}_t)$,
where $\theta_t$ denotes model parameters and $\mathcal{H}_t$ is the interaction history. The underlying structural causal model remains invariant,
$\mathcal{M} = (\mathcal{U}, \mathcal{V}, \mathcal{F}_{\theta_t}, P_{\theta_t}(\mathcal{U}))$,
and only parameters inside a fixed functional class are updated. In this regime, the evolution operator induces updates only in parameters $\theta_t$ while keeping the structural causal model class fixed.

\textit{(2) Fixed structure with intervention semantics (causal world models).}
Causal world models extend observational learning by introducing interventional reasoning via do-calculus, where observational and interventional distributions differ as $P^{\mathrm{obs}}_{\theta}(s_{t+1} \mid s_t, a_t) \neq P_{\theta}(s_{t+1} \mid s_t, do(a_t))$. However, despite enabling causal interpretation of actions, the structural causal model remains fixed over time, meaning $\mathcal{M}_t \equiv \mathcal{M}$ for all $t$.
Thus, interventions are interpreted within a static causal structure, without modifying the underlying mechanisms.

\textit{(3) Self-evolving structural operator (this work).}
In contrast, we extend the causal model itself into a dynamic object by introducing a time-indexed SCM,
$\mathcal{M}_t = (\mathcal{U}, \mathcal{V}, \mathcal{F}_{\theta,t}, P_{\theta,t}(\mathcal{U}))$,
where both structural mechanisms and uncertainty representations are allowed to evolve.
We define a structural evolution operator $\Phi$ that updates the causal world model over time using both the current model state and the accumulated interaction history, which consists of all past state–action–observation tuples.
Importantly, this operator acts on causal mechanisms themselves, meaning that
$(\mathcal{F}_{\theta,t+1}, P_{\theta,t+1}(\mathcal{U})) \neq (\mathcal{F}_{\theta,t}, P_{\theta,t}(\mathcal{U}))$,
which distinguishes structural evolution from purely parametric updates.
Rather than interpreting prior work as a combination of components, we characterize all paradigms by constraints on the admissible evolution operator $\Phi$.
In conventional learning, the induced evolution of $\mathcal{M}_t$ under $\Phi$ reduces to parameter-only updates within a fixed structural class. 
In causal world models, $\Phi$ is restricted to transformations that preserve a fixed SCM while enabling interventional reasoning within the SCM, and in the proposed framework, $\Phi$ is extended to allow modifications of both structural equations and uncertainty distributions.
These regimes can therefore be summarized as different admissible transformation spaces over causal models, rather than different combinations of model components.
Under this formulation, embodied Intelligence is defined as continual causal knowledge construction over the trajectory of evolving causal world models $\{\mathcal{M}_t\}_{t\ge 0}$.


\section{Causal World Models as Internal Generative Mechanisms}

A causal world model is not merely a predictor of future observations, but a structured representation of the latent mechanisms governing environmental dynamics. Conventional world models typically learn observational transition distributions of the form $P_{\theta}(s_{t+1} \mid s_t, a_t)$, which capture statistical regularities in state-action trajectories induced by interaction data collected from agent-environment interactions.

We define a causal world model as a learned approximation of the SCM, as follows:
\begin{equation}
\mathcal{M}' = (\mathcal{U}, \mathcal{V}, \mathcal{F}_{\theta'}, P_{\theta'}(\mathcal{U})), \label{eq:SCM-approx}
\end{equation}
where $\theta'$ is a set of estimated parameters, $\mathcal{F}_{\theta'} = \{ f_{\theta'}, g_{\theta'} \}$ approximates the underlying structural mechanisms governing environmental dynamics, and $\mathcal{M}'$ approximates $\mathcal{M}$.
We further model actions as interventions via an operator, defined as a mapping from the causal world model to an interventional model conditioned on the executed action:
\begin{equation}
do(a_t) : \mathcal{M'} \rightarrow \mathcal{M'}_{a_t}, \label{eq:interv}
\end{equation}

where $a_t$ is an externally imposed assignment, thereby removing the influence of its original causal parents and yielding the corresponding interventional model. Under this formulation, the model predicts interventional outcomes:
\begin{equation}
\hat{s}_{t+1} = f_{\theta'}(s_t, a_t, \hat{u}_t)
\end{equation}
corresponding to the interventional distribution $P_{\theta'}(s_{t+1} \mid s_t, do(a_t))$, which captures how the environment responds under externally imposed action $a_t$, and $\hat{u}_t$ is an estimation of $u_t$.

Beyond forward prediction, causal world models enable counterfactual reasoning over alternative interventions. Given an observed transition $(s_t, a_t, s_{t+1})$, 
the model first infers the underlying exogenous noise $\hat{u}_t$ via an abduction process. It can then evaluate the counterfactual outcome of an alternative action $a'$ via:
\begin{equation}
\hat{s}_{t+1}^{a'} = f_{\theta'}(s_t, a', \hat{u}_t).
\end{equation}

To satisfy the consistency property of structural causal models, which requires the counterfactual prediction to coincide with the factual observation when the intervention matches the observed action ($a' = a_t$), we introduce the following self-consistency objective:

\begin{equation}
\mathcal{L}_{\textrm{cf}} = d\left(s_{t+1}, \hat{s}_{t+1}^{a_t} \right), 
\end{equation}
where $d()$ is a distance metric. Minimizing $\mathcal{L}_{\textrm{cf}}$ encourages the model to accurately align its internal structural mechanisms with the factual environment feedback, ensuring valid causal identifier estimation.

\subsection{Intervention Awareness}

Within the SCM formulation, intervention awareness refers to the ability of an embodied agent to distinguish observational transitions generated under $P_{\theta,t}(s_{t+1} \mid s_t, a_t)$ from interventional dynamics induced by explicit causal manipulations of actions via $do(a_t)$. Unlike purely observational learning, interventional signals reveal causal dependencies encoded in the structural equations $\mathcal{F}_{\theta,t}$ that may remain latent under observational correlations.

This capability enables agents to identify invariant structural mechanisms governing environmental dynamics by isolating the causal effects of actions from confounding observational dependencies. As a result, intervention-aware agents reduce reliance on spurious statistical regularities and achieve improved robustness under distribution shifts and out-of-distribution conditions, where observational correlations may no longer hold.

Intervention awareness enables embodied agents to move beyond passive prediction toward active manipulation of the environment. Rather than solely modeling $P_{\theta,t}(s_{t+1} \mid s_t, a_t)$, agents explicitly query interventional dynamics of the form $P_{\theta,t}(s_{t+1} \mid s_t, do(a_t))$, thereby using the environment as an experimental system to probe underlying causal mechanisms.

More fundamentally, interventions serve as structured queries to the structural causal model, allowing agents to generate informative evidence for causal hypothesis evaluation. Learning therefore becomes an iterative process of causal experimentation, where the structural functions $\mathcal{F}_{\theta,t}$ and the exogenous noise model $P_{\theta,t}(\mathcal{U})$ are refined based on the outcomes of interventions.

Importantly, within the self-evolving cognitive framework, intervention outcomes provide direct feedback signals for the update process $\Phi$. Through interaction history $\mathcal{H}_t$, the agent continuously revises its causal world model $\mathcal{M}_t = (\mathcal{U}, \mathcal{V}, \mathcal{F}_{\theta,t}, P_{\theta,t}(\mathcal{U}))$, enabling progressive refinement of causal identifiability, structural accuracy, and robustness under distribution shifts. This establishes a direct connection between interventional data generation and the self-evolving update process $\Phi$, where interventions serve as the primary source of structural refinement signals for the causal world model.

Ultimately, intervention awareness constitutes a key capability for self-evolving causal cognition. By actively generating interventional evidence rather than passively consuming observational data, embodied agents can continuously refine their causal world models through interaction-driven feedback, enabling robust causal discovery, improved generalization, and increasingly accurate explanatory representations of environmental dynamics.

\subsection{Counterfactual Reasoning}

Causal world models enable counterfactual reasoning over alternative histories and hypothetical futures by querying structural equations under intervened actions and alternative realizations of exogenous variables. Specifically, given the learned SCM $\mathcal{M}_t = (\mathcal{U}, \mathcal{V}, \mathcal{F}_{\theta,t}, P_{\theta,t}(\mathcal{U}))$, counterfactual inference considers hypothetical scenarios generated by interventions on actions via $do(a_t)$ and conditioning on alternative realizations of exogenous variables, allowing embodied agents to evaluate how outcomes would change under different causal circumstances.

Reasoning extends beyond predictive rollouts under a fixed policy and becomes the evaluation of alternative causal trajectories induced by interventions and exogenous variability under fixed structural mechanisms $\mathcal{F}_{\theta,t}$. This enables embodied agents to explore competing causal explanations of observed outcomes by simulating ``what-if'' worlds that are consistent with the learned structural causal model, without requiring physical execution in the real environment.

Such counterfactual reasoning can be interpreted as a form of internal causal experimentation, where the agent leverages the structural equations to generate alternative outcomes under hypothetical interventions. Crucially, this process is not purely passive imagination but is grounded in the current causal world model and is therefore constrained by the learned structural mechanisms encoded in $\mathcal{F}_{\theta,t}$.

Importantly, counterfactual reasoning is performed under the current model instance $\mathcal{M}_t$, while the model itself is continuously updated over time through the self-evolving process $\Phi$. This induces a dynamic sequence of causal world models $\{\mathcal{M}_t\}_{t=0}^{T}$, where each instance refines its structural functions and exogenous distributions based on interaction history, enabling progressively more accurate counterfactual inference as the causal representation improves.

Furthermore, within the self-evolving cognitive framework, counterfactual discrepancies between predicted outcomes under interventions and observed outcomes in the real environment provide feedback signals for causal refinement. Through the update process $\Phi$, the agent revises $\mathcal{F}_{\theta,t}$ and the distribution over exogenous variables $P_{\theta,t}(\mathcal{U})$ to improve counterfactual consistency, thereby enhancing causal identifiability, robustness under distribution shifts, and intervention generalization.
Ultimately, counterfactual reasoning serves as a key computational mechanism for bridging prediction, intervention, and causal discovery within a self-evolving causal system, enabling embodied agents to continuously refine their causal world models through interaction, intervention, and feedback.

\subsection{Mechanism-Level Generalization}

Unlike conventional policy-level generalization, causal world models generalize through invariant structural mechanisms underlying environmental dynamics, as defined by the SCM formulation. Consequently, they remain robust under distribution shifts where superficial statistical patterns change while the underlying causal structure remains invariant. Because the model captures latent causal mechanisms rather than merely fitting observational trajectories, acquired knowledge can naturally transfer across environments with different embodiments, configurations, or interaction conditions. Such mechanism-level generalization is particularly important for embodied intelligence, where both environmental dynamics and embodiment properties may evolve over time. By grounding knowledge in causal structure rather than surface correlations, embodied agents can adapt to novel situations while preserving previously acquired causal understanding.

From an epistemic perspective, mechanism-level generalization enables embodied agents to reuse explanatory causal knowledge beyond the specific contexts in which it was acquired. Rather than treating knowledge as task-specific behavioral patterns, causal world models encode transferable abstractions corresponding to invariant components of the structural equations $\mathcal{F}_{\theta,t}$ and the underlying exogenous distributions $P_{\theta,t}(\mathcal{U})$. This allows learned causal knowledge to remain valid across diverse environments governed by the same or approximately invariant causal mechanisms. As a result, agents can achieve efficient adaptation to unfamiliar tasks and reduce reliance on extensive retraining when environmental conditions change.

Achieving such mechanism-level generalization requires identifying latent variables corresponding to invariant causal factors that remain stable across environments despite variations in surface-level observations. In the structural causal model, endogenous state variables $s_t$ and exogenous variables $u_t \in \mathcal{U}_s$ can be encoded into compact low-dimensional causal representations via causal representation learning methods, including structured variational latent-variable models and contrastive causal representation learning (CRL)~\cite{Scholkopf2021}. 

Within this formulation, mechanism-level generalization corresponds to learning structural functions that are invariant or slowly varying under the self-evolving refinement process $\Phi$, such that the updated model $\mathcal{M}_t = (\mathcal{U}, \mathcal{V}, \mathcal{F}_{\theta,t}, P_{\theta,t}(\mathcal{U}))$ preserves stable causal mechanisms across environments. The learned structural equations $\mathcal{F}_{\theta',t}$ then operate over disentangled causal abstractions rather than raw sensory observations, where $\theta',t$ denotes parameters updated through $\Phi$ based on interaction history, enabling transfer across different embodiments, environments, and interaction conditions while preserving invariant causal mechanisms underlying environmental dynamics.

Ultimately, mechanism-level generalization allows embodied agents to move beyond memorizing environment-specific behaviors toward acquiring reusable causal knowledge. As the causal world model is continually refined through interaction, intervention, and feedback via $\Phi$, the resulting knowledge can be transferred across tasks and environments, supporting continual adaptation and contributing to the emergence of embodied scientific intelligence.

\section{Intervention-Driven Self-Evolving Causal Learning}

We propose a self-evolving cognitive framework composed of three recursively interacting processes: (i) self-observation, (ii) self-evaluation, and (iii) self-updating. Rather than operating solely on external behavioral outcomes, these processes recursively refine an internal causal world model through continual embodied interaction of the agent, enabling simultaneous improvement of perception, reasoning, and decision making.

Formally, the recursive cognitive dynamics are defined as
\begin{equation}
(\mathcal{F}_{\theta',t+1},
P_{\theta',t+1}(\mathcal{U}),
\pi_{\theta',t+1})
=
\Phi(
\mathcal{F}_{\theta',t},
P_{\theta',t}(\mathcal{U}),
\pi_{\theta',t},
\mathcal H_t),
\end{equation}
where $\mathcal H_t$ denotes the accumulated interaction history and $\theta'$ denotes the inference-time or imagination-time parameters of the causal world model used for abduction and counterfactual simulation.
At each iteration, the embodied agent updates both its internal causal world model and behavioral policy through a closed-loop causal learning process consisting of the following steps:

\begin{enumerate}
\item \textbf{Observation:} Collect the empirical interaction trajectory $(s_t, a_t, s_{t+1})$;

\item \textbf{Abduction and prediction:} Infer the exogenous noise $\hat{u}$ conditioned on the factual transition, and reconstruct the predicted state $\hat{s}_{t+1} = f_{\theta',t}(s_t, a_t, \hat{u}_t)$;

\item \textbf{Counterfactual simulation:} Evaluate an alternative hypothetical intervention $a'$, and simulate the counterfactual outcome $\hat{s}_{t+1}^{a'} = f_{\theta',t}(s_t, a', \hat{u}_t)$;

\item \textbf{Causal discrepancy:} Measure structural and mechanism inconsistency at time step $t$:
\begin{equation}
\delta_t = d(s_{t+1}, \hat{s}_{t+1}) + \lambda \cdot d\left(\hat{u}_t, \mathcal{E}_{\theta',t}(\hat{s}_{t+1}^{a'}, a', s_t)\right),
\end{equation}
where $\mathcal{E}_{\theta',t}$ denotes the causal abduction network. The second term acts as a counterfactual invariance regularizer, enforcing that the inferred environmental exogenous stochasticity remains invariant under alternative interventions.

\item \textbf{Model update:} Perform online structural refinement via gradient descent to update the parameters for the next iteration:
\begin{equation}
\theta'_{t+1} \leftarrow \theta'_t - \eta \nabla_{\theta'_t} \delta_t,
\end{equation}
where the policy $\pi_{\theta', t+1}$ is subsequently optimized using the representations refined by $\delta_t$ (e.g., via policy gradient over the updated world model).
\end{enumerate}

This defines a closed-loop causal learning system in which prediction, intervention, and counterfactual reasoning jointly drive representation learning and model refinement.

Importantly, the learning signal is not limited to observational prediction error, but also incorporates discrepancies arising from hypothetical interventions, ensuring that model updates are guided by underlying causal mechanisms rather than purely correlational statistics. Failures within this loop are therefore not merely undesirable outcomes, but informative signals that reveal mismatches between internal causal hypotheses of the agent and the true mechanisms governing the environment. In this sense, failure functions as an implicit diagnostic intervention, exposing hidden dependencies, latent variables, or incorrect structural assumptions embedded in the learned causal world model.

Crucially, self-updating is driven by the full spectrum of causal reasoning processes introduced in previous sections, including intervention-aware data generation, counterfactual simulation, and mechanism-consistent evaluation. Through these processes, the agent evaluates alternative causal trajectories and uses structured causal feedback to refine its internal model. This closes the loop between causal reasoning and learning, enabling the system to evolve not only its behavioral policy but also its internal causal world model and generative understanding of the environment. Learning is no longer a process of passive optimization, but a continual epistemic process in which the agent actively constructs, tests, and revises causal explanations of its environment through embodied interaction. The resulting system can therefore be viewed as a self-evolving cognitive agent capable of continual causal discovery, hypothesis refinement, and autonomous model revision in open-ended environments, providing a foundation for embodied scientific intelligence.

\section{Embodied Scientific Intelligence through Self-Evolving Cognitive Framework}

A major implication of the proposed framework is the emergence of embodied scientific intelligence. Existing embodied systems primarily focus on task execution, manipulation, navigation, or short-horizon decision making. We argue that self-evolving causal world modeling enables a transition from task-oriented behavior toward higher-order scientific reasoning, adaptive experimentation, and continual knowledge acquisition.

Embodied agents no longer operate solely as reactive executors of predefined behaviors, but progressively develop the capacity to interpret observations, infer causal dependencies, generate and evaluate hypotheses, and refine internal causal world models through continual interaction with their environments via the self-evolving operator $\Phi$.

We characterize embodied scientific intelligence through four interconnected levels:

\begin{itemize}
\item \textbf{Execution}: interaction with the environment and execution of actions;
\item \textbf{Inference}: interpretation of observations and causal explanation of environmental outcomes under the learned structural causal model, conditioned on the inference-time parameterization $\theta'$;
\item \textbf{Design}: formulation of intervention strategies and hypothesis generation via interventional queries of the form $do(a_t)$ and counterfactual simulation within the structural causal model $\mathcal{F}_{\theta',t}$;
\item \textbf{Discovery}: identification of novel causal structures through iterative experimentation and model refinement driven by the self-evolving operator $\Phi$.
\end{itemize}

Importantly, these levels should not be viewed as isolated stages, but as mutually reinforcing components of a continuous causal cognitive cycle induced by $\Phi$. Knowledge acquired through interaction and intervention is incorporated into the causal world model of the agent, thereby improving subsequent inference, intervention design, and discovery.

Crucially, at the \textbf{Design} level, agents acquire the ability to actively select interventions and evaluate hypothetical outcomes, both in physical environments and within simulated causal rollouts. These simulations are performed using the structural causal model $\mathcal{F}_{\theta',t}$ conditioned on the inference-time parameterization $\theta'$, enabling structured exploration of causal hypotheses prior to real-world execution and reducing interaction cost while improving the efficiency of knowledge acquisition.

Within the proposed framework, embodied scientific intelligence emerges through a continual $\Phi$-driven epistemic loop operating over the structural causal model:

\[
\begin{aligned}
\text{observation}
&\rightarrow \text{abduction and causal inference via } \theta'
\rightarrow \text{intervention } (do(a_t)) \\
&\rightarrow \text{counterfactual evaluation under } \mathcal{F}_{\theta',t}
\rightarrow \text{outcome estimation}
\rightarrow \text{model revision via } \Phi
\end{aligned}
\]

This loop explicitly incorporates counterfactual simulation as a causal reasoning stage in which hypothetical interventions are evaluated under the learned structural causal model before execution. Through intervention-based and counterfactual reasoning, agents can anticipate, compare, and select among alternative outcomes under different causal assumptions, thereby supporting both decision making and hypothesis evaluation.
Through repeated execution of this $\Phi$-driven loop, the causal world model is continuously revised, expanded, and refined. Knowledge acquisition therefore emerges not as a terminal outcome, but as an ongoing process of self-evolution in which newly acquired causal knowledge recursively improves future reasoning, experimentation, and model construction.

Within this framework, embodiment serves not only as a mechanism for interaction with the environment, but also as a medium for epistemic engagement with the world. Experience is therefore accumulated not merely as behavioral trajectories, but as progressively refined causal knowledge regarding environmental structure, dynamics, and intervention-dependent relationships.
Embodied systems may evolve beyond execution-centric automation toward adaptive cognitive systems capable of continual experimentation, causal reasoning, knowledge acquisition, and self-evolving scientific intelligence through tightly coupled embodied and simulated interaction. The proposed framework is embodiment-agnostic and can be instantiated across both virtual and physical embodied agents that interact with their environments through perception and action.

\section{Causal-Epistemic Interaction via SCM-Based Counterfactual Reasoning}

We define simulated experience as a structured epistemic substrate grounded in the learned structural causal model and conditioned on the inference-time parameterization $\theta'$, through which embodied agents reason about alternative environmental states, hypothetical interventions, and potential outcomes under the self-evolving causal world model $\mathcal{M}_t = (\mathcal{U}, \mathcal{V}, \mathcal{F}_{\theta',t}, P_{\theta',t}(\mathcal{U}))$.
Formally, we represent simulated experience as an epistemically enriched interaction dataset:
\begin{equation}
\mathcal{D}_{\mathrm{sim}} = \{ (s_t, a_t, s_{t+1}, a', \hat{s}_{t+1}^{a'} ) \},
\end{equation}
where $\hat{s}_{t+1}^{a'} = f_{\theta',t}(s_t, a', \hat{u}_t \mid \theta')$ denotes the counterfactual outcome under intervention $do(a')$, obtained by abduction of the exogenous variable $\hat{u}_t$ from the factual transition under the inference-time parameterization $\theta'$.

Rather than viewing simulation as a tool for purely predictive rollout, we interpret simulated experience as internal SCM-based reasoning over intervention-conditioned trajectories. It enables embodied agents to evaluate hypothetical interventions, explore alternative causal pathways, replay failure cases, and refine structural hypotheses without physical execution in the real environment.
Simulated experience corresponds to structured inference over the interventional and counterfactual semantics of the SCM, rather than simple trajectory generation under an observational policy.

We distinguish simulated experience along three primary dimensions:

(i) \textit{Counterfactual exploration}, which enables reasoning over alternative outcomes under different interventions $do(a')$ conditioned on the same inferred exogenous realization $\hat{u}_t$;

(ii) \textit{Interventional simulation}, which evaluates the consequences of explicit causal manipulations within the structural functions $\mathcal{F}_{\theta',t}$;

(iii) \textit{Mechanism-level abstraction}, which supports learning invariant causal mechanisms beyond surface-level observational trajectories.

Together, these dimensions characterize simulated experience as an epistemic extension of the SCM, enabling causal reasoning, hypothesis evaluation, and self-evolving model refinement driven by the operator $\Phi$.

\subsection{Fidelity}

Fidelity refers to the degree to which simulated outcomes preserve the structural consistency of the underlying causal world model with respect to the true environmental SCM. High-fidelity simulation implies that the learned structural functions $\mathcal{F}_{\theta',t}$ and exogenous distribution $P_{\theta',t}(\mathcal{U})$ closely approximate the true causal mechanisms, enabling reliable interventional and counterfactual inference.

However, fidelity should not be interpreted solely as perceptual realism. Lower-fidelity simulation can still provide significant epistemic value by enabling abstraction and mechanism isolation. By simplifying environmental complexity, agents can better disentangle causal structure from observational noise, facilitating hypothesis generation and intervention design under controlled structural variations. Simulated interaction operates across a spectrum of fidelity regimes, each supporting different levels of causal abstraction, mechanism discovery, and intervention-driven learning.

\subsection{Counterfactuality}

Counterfactuality refers to the capacity to evaluate alternative hypothetical outcomes under interventions and alternative realizations of exogenous variables within the SCM.
Formally, this corresponds to evaluating interventional distributions of the form:
\begin{equation}
P_{\theta',t}(s_{t+1} \mid s_t, do(a')),
\end{equation}
which characterizes the structural response of the environment under externally imposed interventions.
This transforms simulation into a structured causal reasoning space, enabling agents to evaluate:
(i) future contingencies under alternative interventions,
(ii) counterfactual historical outcomes under modified actions,
and (iii) alternative environmental realizations under different exogenous configurations.

Such reasoning supports not only planning and prediction, but also causal attribution, failure diagnosis, and structural hypothesis evaluation. Importantly, counterfactual discrepancies between simulated outcomes and factual observations provide learning signals for refining $\mathcal{F}_{\theta',t}$ and $P_{\theta',t}(\mathcal{U})$, thereby integrating simulation directly into the self-evolving update process $\Phi$.

\subsection{Agency}

Agency refers to the degree to which simulated outcomes are conditioned on the ability of an embodied agent to actively select interventions and generate structured queries over the SCM.
High-agency simulation enables agents not only to evaluate passive trajectories, but also to design interventions $do(a)$, explore their consequences, and optimize decision strategies based on intervention-conditioned outcomes. This supports active experimentation, hypothesis testing, and causal discovery in both real and simulated environments.

In contrast, low-agency simulation corresponds to passive rollout prediction without meaningful intervention control. Agency transforms simulation from a predictive tool into an interactive causal experimentation environment, where agents actively probe structural mechanisms, test hypotheses, and refine their internal causal world models through intervention-driven feedback. As a whole, fidelity, counterfactuality, and agency characterize simulated experience not merely as an auxiliary computational mechanism, but as a structured epistemic extension of the self-evolving causal world model, enabling embodied scientific reasoning and continual causal learning.

\section{Intervention-Driven Causal-Epistemic Benchmarking for Embodied Scientific Intelligence}

Traditional robotic benchmarks centered primarily on task completion rates or trajectory imitation accuracy are insufficient for evaluating self-evolving embodied systems. Such metrics typically measure short-horizon behavioral performance under relatively fixed environmental conditions, but fail to capture the capacity for causal reasoning, adaptive model revision, and continual epistemic accumulation.

We therefore argue that embodied intelligence systems should be evaluated not only by execution success, but also by their ability to construct, maintain, and refine internal causal world models through interaction. Evaluation can be formalized around a set of causal and epistemic criteria grounded in intervention-based reasoning. The first four metrics admit explicit operational formulations, whereas metrics (5) and (6) characterize higher-level epistemic properties that are instantiated in task-specific implementations. We define key evaluation dimensions as follows.

\textbf{(1) Causal intervention robustness}

Causal intervention robustness measures the ability of the system to maintain accurate predictive performance under structural manipulations of actions. Formally, we evaluate
\begin{equation}
\mathcal{R}_{\textrm{int}} =
\mathbb{E}_{a_t \sim \mathcal{A}_{\textrm{int}}}
\left[
\mathbb{I}\big( d(\hat{s}_{t+1}, s_{t+1}) < \epsilon \big)
\right],
\end{equation}
where $\hat{s}_{t+1} \sim P_{\theta'}(s_{t+1} \mid s_t, do(a_t))$ denotes the predicted interventional outcome under an externally imposed action, and $s_{t+1}$ is the corresponding ground-truth environmental response.

\textbf{(2) Counterfactual accuracy}

Counterfactual accuracy evaluates the ability to predict alternative histories under counter-to-fact interventions. Utilizing simulator-derived parallel worlds to obtain ground-truth counterfactual states $s_{t+1}^{a'}$, we define
\begin{equation}
\mathcal{E}_{\textrm{cf}} =
\mathbb{E}_{a' \sim \mathcal{A}_{\textrm{int}}}
\left[
d\big(s_{t+1}^{a'}, \hat{s}_{t+1}^{a'}\big)
\right],
\end{equation}
which measures whether the system correctly models how environmental outcomes diverge under hypothetical causal interventions.

\textbf{(3) Failure utilization}

Failure utilization measures how effectively a system converts unexpected outcomes into improvements in its internal causal world model:
\begin{equation}
\mathcal{U}_{\textrm{fail}} =
\frac{\Delta \mathcal{M}_{\text{causal}}}{N_{\text{fail}}}.
\end{equation}
This quantifies whether failure signals are properly integrated as causal feedback for epistemic refinement.

\textbf{(4) Long-horizon causal consistency}

Long-horizon causal consistency evaluates whether the updated internal mechanisms preserve structural invariance over extended sequences, avoiding representational drift:
\begin{equation}
\mathcal{C}_{\textrm{long}} =
\sum_t D_{\mathrm{KL}}\Big(
P_{\theta', t}(\mathcal{U}) \,\|\, P_{\theta', t+k}(\mathcal{U})
\Big).
\end{equation}
This measures the divergence between exogenous uncertainty estimates of the self-evolving model across temporal checkpoints.

\textbf{(5) Self-modification stability}

Self-modification stability evaluates whether continual self-updating improves or destabilizes the learned causal structure, capturing the trade-off between adaptability and representational stability in self-evolving systems.

\textbf{(6) Cross-platform transfer robustness}

Cross-platform transfer robustness measures whether learned causal representations remain invariant across different embodiments, sensor configurations, and environmental conditions. Unlike standard transfer learning metrics, this evaluates structural generalization of causal knowledge rather than policy-level performance.

\medskip
Benchmarking embodied intelligence shifts from measuring static task performance toward evaluating causal understanding, intervention-aware reasoning, and continual epistemic model refinement in open-ended environments.
All evaluation signals are assumed to be integrated into the self-evolving update operator $\Phi$, thereby inducing a feedback-driven refinement of $\mathcal{F}_{\theta',t}$, $P_{\theta',t}(\mathcal{U})$, and the policy $\pi_{\theta',t}$.

\section{From Predictive Intelligence to Self-Evolving Epistemic Intelligence}

Most existing embodied AI systems are primarily optimized for predictive control, focusing on forecasting future states, imitating expert trajectories, or maximizing task-specific rewards. While effective in constrained environments, these approaches predominantly capture observational statistical regularities in states and actions, with limited explicit mechanisms for causal reasoning, intervention modeling, or continual revision of internal knowledge under environmental feedback. In contrast, scientific inquiry is fundamentally epistemic in nature. Its objective extends beyond generating effective actions to constructing, testing, and refining explanatory models of the environment. Rather than operating solely on observable trajectories, epistemic intelligence seeks to uncover the latent causal mechanisms that generate observations and govern environmental dynamics.

We define epistemic intelligence as an embodied learning paradigm in which agents maintain structured causal world models of their environment, represented as a structural causal model, and continuously refine these models through interaction, intervention, counterfactual reasoning, and environmental feedback. Unlike predictive intelligence, which emphasizes mapping observations to future outcomes, epistemic intelligence focuses on identifying, validating, and updating the underlying generative causal structure responsible for those outcomes. 
The evolution of embodied intelligence moves beyond behavior optimization toward causal understanding. Learning is no longer viewed merely as policy optimization, but as an iterative process of causal hypothesis generation, intervention-driven experimentation, causal inference, and continual model refinement governed by a self-evolving update operator $\Phi$.

To realize this transition, embodied agents must maintain explicit representations of latent environmental causal mechanisms encoded in their internal SCM, evaluate hypothetical interventions through $do(\cdot)$-based reasoning and counterfactual inference, and continuously revise these models through closed-loop interaction with the environment. Such agents function not merely as adaptive decision-makers, but as self-improving epistemic systems whose causal world models are continuously updated via $\Phi$. More importantly, interaction should be reinterpreted not only as a means of trajectory optimization, but as a mechanism for causal hypothesis generation, intervention-driven experimentation, and continual epistemic model refinement. Experience therefore serves not only as behavioral data for policy learning, but also as epistemic evidence for updating $\mathcal{F}_{\theta',t}$ and $P_{\theta',t}(\mathcal{U})$ within the SCM.

By integrating perception, intervention, and counterfactual reasoning within a unified SCM-based and $\Phi$-driven framework, embodied agents progressively refine their internal causal world models while simultaneously improving decision-making capability in open-ended environments. This transition from predictive intelligence to epistemic intelligence provides a conceptual foundation for embodied scientific intelligence, enabling agents not only to acquire knowledge autonomously, but also to construct explanatory causal models and uncover the latent mechanisms governing environmental dynamics through continual interaction with the world.

\section{Discussion}

While the proposed framework offers a unified foundation for embodied scientific intelligence, its practical realization poses several important research challenges, particularly with respect to continual self-evolution and adaptive causal knowledge acquisition. Addressing these challenges will be essential for advancing embodied scientific intelligence toward practical real-world systems.

\begin{itemize}

\item \textbf{Continual causal self-evolution.}
How to continually update internal causal world models while preserving invariant causal mechanisms and avoiding representational drift remains a fundamental challenge. Developing principled mechanisms that balance structural stability with continual adaptation during long-term self-evolution is an important research direction.

\item \textbf{Failure-driven knowledge acquisition.}
Unexpected observations, inaccurate predictions, and unsuccessful interventions should be interpreted not merely as optimization errors but as epistemic evidence for revising causal hypotheses and discovering previously unknown causal mechanisms. Developing effective mechanisms for failure-driven causal learning remains an important open research problem.

\item \textbf{Long-horizon causal memory and knowledge consolidation.}
Embodied scientific intelligence requires memory architectures capable of accumulating, abstracting, consolidating, and retrieving causal knowledge over long interaction histories while mitigating catastrophic forgetting and preserving structurally meaningful causal representations.

\item \textbf{Safe intervention and autonomous experimentation.}
As embodied agents become capable of actively designing and executing interventions, integrating safety constraints into intervention-driven reasoning and autonomous experimentation becomes increasingly important for enabling reliable deployment in open-world environments.

\item \textbf{Emergence of higher-order scientific cognition.}
An important open question is how continual causal self-evolution gives rise to higher-order cognitive abilities, including abstraction, analogical reasoning, systematic reasoning, scientific creativity, and autonomous discovery. Understanding the computational principles underlying these capabilities may provide new directions for extending the proposed framework toward more general embodied scientific intelligence.

\item \textbf{Multi-agent causal attribution and self/external intervention disentanglement.}
In open-ended embodied environments, state transitions may be jointly influenced by the focal agent, other agents (e.g., humans or robots), and unobserved environmental factors. This raises a fundamental causal attribution problem: determining whether an observed transition is induced by interventions executed by the agent or by exogenous disturbances. Addressing this challenge may require extending the proposed framework to interacting structural causal models, where multiple agents exert overlapping yet distinguishable intervention effects on a shared environment. A key research direction is to develop principled mechanisms for disentangling self- and other-induced interventions as well as attributing causal responsibility in such interactive settings.
\end{itemize}

These challenges together outline a research roadmap for advancing the proposed self-evolving cognitive framework via causal world modeling toward practical embodied scientific intelligence capable of continual causal discovery, intervention-driven experimentation, adaptive knowledge accumulation, and autonomous scientific reasoning in open-ended environments.

\section{Conclusion}

In this paper, we propose a self-evolving cognitive framework via causal world modeling for embodied scientific intelligence and outline a corresponding intervention-driven causal-epistemic evaluation grounded in causal understanding, intervention-aware reasoning, continual model refinement, and long-term knowledge accumulation. Rather than viewing embodied agents as task-oriented controllers that optimize only performance or reward signals, we advocate embodied systems that continually construct, maintain, and refine structured causal representations of their environments through interaction.

This work presents a unified theoretical framework that formalizes embodied intelligence as an epistemic process grounded in structural causal models. The proposed framework integrates causal world modeling, intervention-driven causal reasoning, and continual cognitive refinement into a unified cognitive process in which interaction is interpreted not only as a mechanism for behavioral optimization, but also as a source of evidence for causal hypothesis generation, intervention-driven experimentation, and continual refinement of internal causal world models. We instantiate this framework through a time-evolving structural causal model formulation together with a self-evolution operator that enables the progressive adaptation of structural mechanisms and exogenous uncertainty representations.

Moreover, this work highlights a transition from predictive embodied intelligence toward epistemic embodied intelligence. Rather than focusing exclusively on short-horizon action optimization, future embodied systems are expected to increasingly emphasize the continual construction, refinement, and utilization of causal world models for reasoning, decision-making, and knowledge acquisition in open-ended environments. Embodied scientific intelligence emerges not as a specialized capability, but as a natural consequence of self-evolving causal cognition grounded in continual interaction with the environment.


\bibliographystyle{ACM-Reference-Format}
\bibliography{sample-base}


\end{document}